\documentclass[10pt,twocolumn,letterpaper]{article}

\usepackage{sty}
\usepackage{times}
\usepackage{epsfig}
\usepackage{graphicx}
\usepackage{amsmath}
\usepackage{amssymb}
\usepackage{multirow}
\usepackage{booktabs} 
\usepackage{tabularx}
\usepackage{caption}
\usepackage{subcaption}

\usepackage{pifont}

 \usepackage{xcolor}
\usepackage{url}

\usepackage{csquotes}
\usepackage{float}
\usepackage{bm}

\usepackage{amsfonts}       
\usepackage{nicefrac}       
\usepackage{microtype}      

\usepackage{soul}
\usepackage{rotating}

\newcolumntype{?}[0]{!{\vrule width 0.05mm}}

\usepackage{dsfont}

\newcommand\blfootnote[1]{%
  \begingroup
  \renewcommand\thefootnote{}\footnote{#1}%
  \addtocounter{footnote}{-1}%
  \endgroup
}

\usepackage[breaklinks=true,bookmarks=false]{hyperref}

\vcvcfinalcopy 

\begin{document}

\title{Hybrid Sample Synthesis-based Debiasing of Classifier in Limited Data Setting}

\author{
Piyush Arora$^{\star}$, \hspace{0.5cm} Pratik Mazumder$^{\star,\copyright}$\\
Indian Institute of Technology Jodhpur, India \hspace{0.5cm}\\
{\tt\small arora.8@iitj.ac.in, pratikm@iitj.ac.in}
}

\maketitle

\ifvcvcfinal\thispagestyle{empty}\fi

\begin{abstract} 
Deep learning models are known to suffer from the problem of bias, and researchers have been exploring methods to address this issue. However, most of these methods require prior knowledge of the bias and are not always practical. In this paper, we focus on a more practical setting with no prior information about the bias. Generally, in this setting, there are a large number of bias-aligned samples that cause the model to produce biased predictions and a few bias-conflicting samples that do not conform to the bias. If the training data is limited, the influence of the bias-aligned samples may become even stronger on the model predictions, and we experimentally demonstrate that existing debiasing techniques suffer severely in such cases. In this paper, we examine the effects of unknown bias in small dataset regimes and present a novel approach to mitigate this issue. The proposed approach directly addresses the issue of the extremely low occurrence of bias-conflicting samples in limited data settings through the synthesis of hybrid samples that can be used to reduce the effect of bias. We perform extensive experiments on several benchmark datasets and experimentally demonstrate the effectiveness of our proposed approach in addressing any unknown bias in the presence of limited data. Specifically, our approach outperforms the vanilla, LfF, LDD, and DebiAN debiasing methods by absolute margins of 10.39\%, 9.08\%, 8.07\%, and 9.67\% when only 10\% of the Corrupted CIFAR-10 Type 1 dataset is available with a bias-conflicting sample ratio of 0.05.
\end{abstract}

\section{Introduction} 
\blfootnote{$^\star$ All the authors contributed equally.}\blfootnote{$\copyright$ Corresponding author} Deep neural networks are renowned for their state-of-the-art performance in several real-world tasks that were previously limited by human capability~\cite{action_2021_ICCV,styleGAN}. In order to perform a particular task, deep neural networks are generally trained on large volumes of labeled data points. However, the training data often contains biases of different types that get introduced into the training data due to issues with the data collection devices/strategies, biases of the annotator, and other issues. Deep learning models are extremely susceptible to the biases present in the datasets \cite{bias_corr, bias_corr2}, and training them on such datasets causes the models to make biased predictions and suffer from degraded performance when evaluated on data that contradicts the learned biases (bias-conflicting samples). For example, when training a deep learning model to classify between a dog and a bird, there is a possibility that the model learns biased features, such as the background color of the image. In such a case, if a majority of the training samples of birds have a blue sky as the background, the model may learn that the background of a bird image has to be a blue sky, while that of a dog image can vary. Samples that conform with such biases in the dataset (bias-aligned samples) may cause the model to rely on superficial features rather than intrinsic features, such as a bird's beak, to perform the classification task. Such biases in deep learning models can have significant implications in real-world applications and warrant thorough investigation and mitigation. In this work, we address the problem of debiasing deep learning models while training on limited training data without access to the type of bias that affects the training data.

Researchers have been working on techniques to reduce the effect of bias in deep learning models. A majority of prior works \cite{rebias,hex,learning_not_to_learn} that debias deep neural networks rely on incorporating prior knowledge of the bias present in the data. However, this presents several practical limitations, including the infeasibility of gathering labeled data indicating the bias present in it. Further, such debiasing methods are not able to address unspecified biases present in the data. In this work, we focus on a more practical and challenging task that entails addressing bias without prior knowledge. In addressing such biases, some methods~\cite{lff,ldd} deliberately amplify the biased nature of a network while attempting to debias the training of a second network by making it more focused on learning samples that run counter to the prejudice/bias (bias-conflicting samples) of the biased network (loss reweighting). Loss reweighting-based methods attempt to debias the model by forcing it to give more importance to learning the bias-conflicting samples. The authors in \cite{biaswap} propose an image translation-based debiasing method that augments realistic bias-conflicting samples for promoting debiased representations. 

Although previous works on the debiasing of deep neural networks without prior knowledge of bias have shown good progress, these methods generally assume that the model has access to a large quantity of labeled data. This may not be practically feasible in many real-world scenarios where the data is limited due to privacy, connectivity, or other reasons. For example, in order to ensure privacy, models may be trained on local devices without transferring personal data to servers, and local data will usually be limited and may most likely suffer from unwanted biases. Furthermore, in many real-world scenarios, obtaining labeled data can be costly, time-consuming, or challenging to obtain. Additionally, due to the limited availability of training data, the effect of bias on the model prediction may be even more pronounced. Infact, our experimental results in Sec.~\ref{sec:abldata} indicate that the performance of an existing debiasing approach degrades significantly when the model is trained on limited training data. Hence, learning to debias models in a limited data setting is an important problem to address, and we focus on this issue in this work.

In the limited data setting, the model is extremely vulnerable to overfitting due to the relatively higher quantity of bias-aligned samples, due to which the model predictions are heavily biased. The debiasing approaches generally rely on making the model better learn the bias-conflicting samples by increasing the weights of the loss incurred for such samples. However, in the limited data setting, this may also lead to overfitting to the very few bias-conflicting samples. We propose a novel approach to debias deep neural networks such that the effect of biases on the model predictions is reduced even in a limited data setting. We first propose an approach to identify the most likely bias-conflicting samples in the batch. Since we want the model to focus more on these samples but not overfit on them, we combine the content of these bias-conflicting samples with other bias-aligned samples from the same class. This leads to the synthesis of hybrid samples that retain the properties of the bias-conflicting samples to reduce the effect of bias-aligned samples on the model predictions. Due to this synthesis process, the hybrid samples also contain sufficient diversity to prevent the debiasing process from making the model overfit to the few bias-conflicting samples when we increase the focus of the model on bias-conflicting samples through loss reweighting. To the best of our knowledge, no other debiasing method uses such an approach to reduce the effect of biases on the model predictions in the limited data setting. Please note that we experimentally evaluate other debiasing methods that also use reweighting in some other ways and observe that these methods still perform poorly in this setting. We perform experiments on reduced/limited data versions of multiple benchmark datasets, such as Colored MNIST, Corrupted CIFAR, BFFHQ, and experimentally demonstrate that our proposed method significantly outperforms existing debiasing approaches on multiple datasets, in the small dataset regime. 

The following are the contributions of this work:
\begin{itemize}
    \item We experimentally demonstrate that the performance of existing debiasing methods for training deep learning models on data with unknown bias drops significantly in the presence of limited training data.
    \item We propose a novel approach for debiasing deep learning models that can significantly reduce the effect of unknown biases on the model predictions in the limited training data setting.
    \item We also experimentally demonstrate that our proposed approach significantly outperforms debiasing methods even in a setting that does not suffer from data scarcity.
    
\end{itemize}

\section{Related Works}

Studies on debiasing \cite{lff,ldd} in deep neural networks have shown that biases influence the learning process when they are easy to learn and act as shortcuts \cite{shortcut}.  Debiasing a model involves training a model in such a manner that the model predictions do not suffer from the biases present in the training dataset.  Researchers have focused on two broad settings for this problem: one in which the bias is known~\cite{rebias,darlow2020latent,huang2020self} and the other, more realistic setting in which the bias is not known~\cite{lff,ldd,debian}.

\textbf{Debiasing Known Bias.} The authors in \cite{rebias} utilize the Hilbert-Schmidt Independence Criterion in order to ensure that the target model is statistically independent from a set of models that encode the bias. The work in \cite{hex} trains a debiased model with a biased model in an adversarial manner using a texture-oriented network. The approach proposed in \cite{learning_not_to_learn} minimizes the mutual information between feature embedding and the bias in the data. This is achieved by adversarially training a network to predict the bias distribution against the feature extraction network. The work in \cite{easy_way} trains a naive model and then a robust model, such that the robust model learns other generalizable patterns that reduce the effect of bias on the model predictions. The authors in \cite{MAZUMDER2022108449} propose to utilize the bias attribute information to train the feature extractor in such a way that the features of images from the same class are close to each other even if they have different values for the bias attributes and the features of images from different classes are far away from each other even if they have the same values for the bias attributes.

\textbf{Debiasing Unknown Bias.} Some research works have also looked into debiasing a deep learning model trained on data containing an unknown bias. This is a much more practical and real-world setting. The authors in \cite{tofu} experimentally showed that knowledge transfer can improve the robustness of the target model if the original and the target tasks suffer from similar biases. The work in \cite{learning2split} learns to split the dataset in such a way that training the model on one split does not make it generalizable enough for the other split, which indicates the possibility of bias. The work in \cite{lff} intentionally makes a network more biased using custom loss and uses the biased model to reweight loss for different training samples depending upon how hard it is for the biased model to learn it, indicating bias-conflicting samples. The authors in \cite{ldd} propose to improve the representation of the samples by performing a swapping operation of the features from the biased and debiased networks during the training process. The authors in \cite{debian} propose a method that alternatively identifies the bias in the data and performs de-biasing through reweighting of the losses.

Although existing debiasing approaches have shown state-of-the-art results in dealing with unknown bias in the data, their performance in a more realistic environment involving small and biased datasets is yet to be explored. Our experimental results show that such methodologies face a significant reduction in performance over small and biased datasets and perform very similarly to the naive vanilla baseline model. In contrast, our proposed approach significantly outperforms these approaches and the baseline model in the limited data settings.

\section{Proposed Approach}\label{sec:approach}

\subsection{Problem Setting}

In this work, we consider the problem of de-biasing deep learning models trained on limited training data suffering from an unknown bias. This experimental setup involves a training dataset $D$ containing limited training data obtained by randomly sampling ``$p$" percent training data for each class of a biased training dataset. The training data points usually contain several attributes. Some attributes are called bias attributes since they are additional attributes that introduce some type of bias in the dataset. On the other hand, some attributes are called intrinsic attributes, which determine the class the data point belongs to. In this setting, there is no information about the type of bias that is affecting the training data. The ``bias-aligned samples'' refers to samples that exhibit a high correlation between bias attributes and target labels. On the other hand, ``bias-conflicting samples'' refer to samples that occur less frequently and are not correlated with the bias commonly associated with that type of sample in the dataset. The training data contains bias-aligned and bias-conflicting samples, and the bias-conflicting ratio ``$\sigma$" represents the ratio of bias-conflicting samples in the training data. A deep learning model trained naively on this limited training data containing bias, will produce highly biased predictions and perform very poorly on the test data. Therefore, the objective of any approach in this setting is to reduce the effect of bias on the model predictions when the model is trained on the limited training data containing bias without any access to prior knowledge about the type of bias that affects the data.

\begin{figure*}
\begin{center}

\includegraphics[width=0.8\textwidth]{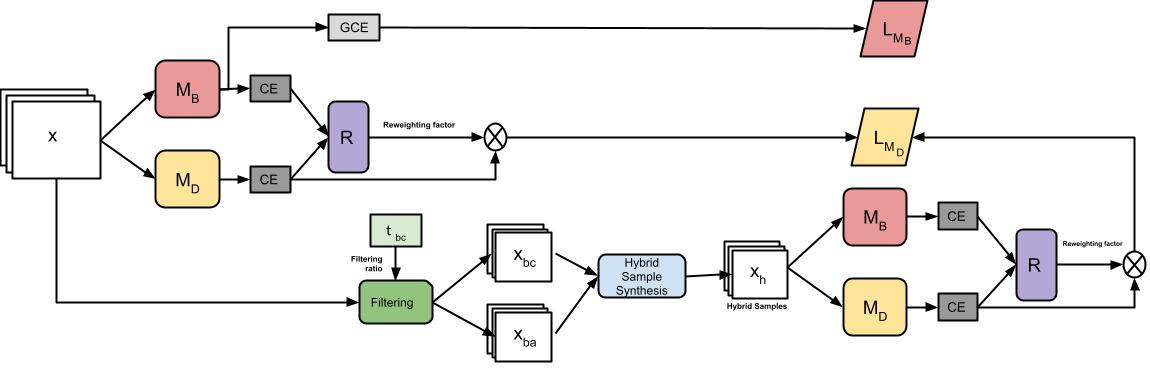}
\end{center}
   \caption{In the proposed approach, we train two models $M_B$ and $M_D$ in parallel. $M_B$ is trained to be biased, whereas, $M_D$ is trained to less biased. $M_B$ is only trained using a generalized cross-entropy based loss $L_{M_B}$ defined in Eq.~\ref{eq:biasloss}. A reweighting factor $R(x)$ is computed for each training sample using Eq.~\ref{eq:reweight} similar to other debiasing methods. In order to deal with the limited data setting, we filter out likely bias-conflicting samples $x_{bc}$ using the procedure described in Sec.~\ref{sec:hybridsample} and combine them with the likely bias-aligned samples $x_{ba}$ to obtain hybrid samples $x_{h}$. We also obtain the reweighting factors $R(x_h)$ for the hybrid samples and train $M_D$ using using a composite loss $L_{M_D}$ defined in Eq.~\ref{eq:debiasloss}. After training, we discard $M_B$ and use $M_D$ for evaluation.}
   \vspace{-15pt}
\label{fig:model}
\end{figure*}

\subsection{Dual Model Training}

In our approach, we train a pair of models ($M_B$,$M_D$) that have similar architectures, in parallel. This follows the idea used by popular debiasing methods~\cite{ldd,lff}, that train $M_B$ to be highly biased and train $M_D$ (target model) to focus more on likely bias-conflicting samples which $M_B$ finds difficult to learn due to the incorporated bias. 

In our approach, $M_B$ is trained to match the erroneous correlations between the bias attributes and the class label and thereby make biased predictions. The bias attributes often make the model learn unwanted correlations and erroneous shortcuts to optimize the loss functions. Consequently, the bias-aligned samples are easier to learn for the model due to the bias attributes present in them. Therefore, in order to make $M_B$ more biased, the model should give more importance to training data points that are easier to learn, i.e., the bias-aligned data points in this case. This can be achieved using the generalized cross entropy (GCE)~\cite{gce} loss that gives more importance to training data points for which there is a high similarity between the model predictions and the label, which is the case with bias-aligned data points in this setting. The $M_D$ model is trained in parallel with $M_B$ but the GCE loss is not used to avoid making $M_D$ more biased. Since $M_B$ will be prone to making more biased predictions and will give a lower loss for bias-aligned data points, we will identify training data points that are difficult for $M_B$ to learn and give more importance to these data points while training $M_D$. This is a popular approach in debiasing methods and the reweighting factor used for changing the importance of the training data points in the training process is obtained as mentioned in Eq.~\ref{eq:reweight}. The obtained reweighting factor is used to reweight the losses incurred by $M_D$ for the training data (see Fig.~\ref{fig:model}).

\begin{equation}\label{eq:reweight}
    R(x) = \frac{L_B(x)}{L_B(x) + L_D(x)}
\end{equation}

Where, $L_B(x)$ and $L_D(x)$ refer to the losses incurred by the models $M_B$ and $M_D$, respectively, for the training data point $x$ with label $y$. Please note that $L_B(x)$ and $L_D(x)$ both use the cross-entropy loss ($L_B(x) = \texttt{CE}(M_B(x),y)$, $L_D(x) = \texttt{CE}(M_D(x),y)$) for computing the reweighting factor. $R(x)$ refers to the reweighting factor for $x$.

\subsection{Synthesizing and Using Hybrid Bias-Conflicting Samples}\label{sec:hybridsample}
As described in Sec.~\ref{sec:intro}, in the limited data setting, the model is even more prone to overfitting to the bias-aligned samples. Similarly, using loss reweighting to give more importance to the very few bias-conflicting samples present in the limited data setting, may also lead to overfitting and poor performance. We propose to address this problem by introducing new hybrid training data points using the existing few bias-conflicting samples to reduce the impact of bias-aligned samples while maintaining sufficient diversity in these new data points to prevent overfitting on the bias-conflicting samples. However, in this setting, the bias-conflicting samples are not known. Instead, we propose to utilize the reweighting factors to identify training data points that are highly likely to be bias-conflicting. This is based on the idea that the training samples for which the reweighting factor is very high, are the ones that the biased model $M_B$ finds very difficult to learn, and therefore, these samples must be in conflict with the bias learned by $M_B$. We propose to identify a filtering ratio ($t_{bc}$) for the training data points in a batch for this process. Assuming a batch of $N$ samples, each with an associated reweighting factor $w_i$, we select the top $N*(1-t_{bc})$ samples with the highest reweighting factor as the most likely bias-conflicting samples. The rest of the samples are considered as likely bias-aligned samples. 

After obtaining the likely bias-conflicting samples, we synthesize hybrid samples by combining these samples with the likely bias-aligned samples from the same class. This has three benefits, as mentioned earlier. First of all, this increases the number of training samples that contain the information of the bias-conflicting samples. Secondly, this process also ensures sufficient diversity in the hybrid samples, thereby preventing the model from overfitting to the few bias-conflicting samples when $M_D$ is trained with more importance for the bias-conflicting samples. We also obtain the reweighting factors for the hybrid samples using Eq.~\ref{eq:reweight}, so that if the hybrid sample has more influence of the bias attribute, then its importance in the training process will be lowered using the reweighting factor. The $M_D$ is trained using both the regular training samples and the hybrid samples with their corresponding reweighting factors (see Fig.~\ref{fig:model}).

\begin{equation}\label{eq:mixup}
    x_h = \alpha \ast x_{bc} + (1-\alpha) \ast x_{ba}
\end{equation}

\begin{equation}\label{eq:rwmixup}
     R(x_h) = \frac{L_B(x_h)}{L_B(x_h) + L_D(x_h)}
\end{equation}

Where, $x_{bc}$ and $x_{ba}$ are the likely bias-conflicting samples and the likely bias-aligned samples, respectively, identified using the filtering ratio $t_{bc}$ on their reweighting factors, with labels $y_{bc}$ and $y_{ba}$, respectively, such that $y_{bc} = y_{ba}$. $\alpha$ is a hyperparameter used to control the contribution of the samples while synthesizing the hybrid samples. $x_h$ refers to the synthesized hybrid sample with the label $y_h = y_{bc} = y_{ba}$. $R(x_h)$ refers to the reweighting factor for $x_h$. $L_B(x_h)$ and $L_D(x_h)$ refer to the loss incurred by the models $M_B$ and $M_D$, respectively, for $x_h$.

To summarize, the $M_B$ model is trained using loss $L_{M_B}$ shown in Eq.~\ref{eq:biasloss}. Please note that $M_B$ is trained using the generalized cross-entropy loss, as mentioned earlier. The $M_D$ model is trained using the loss $L_{M_D}$ shown in Eq.~\ref{eq:debiasloss}. 

\begin{equation}\label{eq:biasloss}
     L_{M_B} = \Sigma_x\ \  \texttt{GCE}(M_B(x),y)
\end{equation}

\begin{multline}\label{eq:debiasloss}
     L_{M_D} = \Sigma_x\ \  \texttt{CE}(M_D(x),y) * R(x) + \\\beta \ast \Sigma_{x_h}\ \  \texttt{CE}(M_D(x_h),y_h) * R(x_h)
\end{multline}

\section{Experiments}

\subsection{Datasets and Implementation Details}

In our experiments, we utilized the reduced/limited data versions of the following datasets: Colored MNIST, Corrupted CIFAR-10, BFFHQ. These datasets are the benchmark datasets to evaluate debiasing techniques. Colored MNIST \cite{lff} is a variant of the original MNIST dataset \cite{mnist} with color bias. Each digit in Colored MNIST is associated with a specific foreground color. Corrupted CIFAR-10, consists of ten different types of texture biases applied to the CIFAR-10 dataset \cite{cifar}. Corrupted CIFAR-10 Type 0 and Corrupted CIFAR-10 Type 1 differ in the type of biases present. The Biased FFHQ (BFFHQ) dataset was derived from the FFHQ dataset \cite{ffhq} by the authors of \cite{ldd}, which consists of real-world human face images labeled with various facial attributes. Please refer to the appendix for further details on the datasets.

For the reduced Colored MNIST dataset, we use $p=5\%$ of the training data, resulting in 3000 training samples, and report the results on different bias-conflicting ratios ($\sigma$), i.e., $0.01, 0.02$ and $0.05$. For the reduced Corrupted CIFAR-10 Type 0 and Type 1 datasets, we set $p$ as $10\%$, resulting in 5000 training samples, and report the results on $\sigma \in \{0.01,0.02,0.05\}$. For the reduced BFFHQ dataset, we set $p$ as $10\%$, resulting in $2000$ training samples, and report results on $\sigma=0.05$. For Colored MNIST and Corrupted CIFAR-10, we use the test data mentioned in \cite{lff}. The test data for the Colored MNIST and Corrupted CIFAR-10 datasets consists of $10000$ samples with $9000$ bias-conflicting samples. For the BFFHQ dataset, we use the validation and test datasets described in \cite{ldd}, which contains $1000$ test images. The bias-conflicting test set for BFFHQ excludes the bias-aligned samples from the unbiased test set. The hyper-parameters $\alpha$, $\beta$ and $t_{bc}$ are identified using ablation experiments. We use $\alpha=0.5$, $\beta=0.2$, and $t_{bc}=0.9$ for Corrupted CIFAR-10, $\alpha=0.9$, $\beta=1.0$, $t_{bc}=0.95$ for Colored MNIST and $\alpha=0.5$, $\beta=0.1$, and $t_{bc}=0.9$ for BFFHQ. We report the accuracy averaged over three runs for different seeds. In our experiments, we utilize the ResNet18 \cite{ResNets} model without pre-trained weights for the Corrupted CIFAR-10 and BFFHQ datasets and a simple MLP model for the CMNIST similar to \cite{lff}. In our experiments, we employ the Adam optimizer \cite{adam}. For our method, we pre-train the model to predict the angle of rotation of the image by assigning pseudo labels (for CIFAR-10, BFFHQ). Please refer to the appendix for further details. Our experimental results are presented for three debiasing methods, namely "LfF" \cite{lff}, "LDD" \cite{ldd}, and "DebiAN" \cite{debian}, as well as the baseline vanilla model that is trained without any debiasing techniques.

\begin{table*}[t]
\centering
\begin{tabular}{cccccc}
\toprule
$\sigma$ & Vanilla & LfF & LDD & DebiAN & Ours \\
\midrule
$0.01$ & 39.47 \scriptsize{$\pm$} \scriptsize{2.50} & 43.66 \scriptsize{$\pm$} \scriptsize{3.48} & 42.34 \scriptsize{$\pm$} \scriptsize{1.82} & 35.86 \scriptsize{$\pm$} \scriptsize{2.89} & \textbf{48.92} \scriptsize{$\pm$} \scriptsize{4.32} \\ 
$0.02$ & 43.66 \scriptsize{$\pm$} \scriptsize{1.13} & 56.58 \scriptsize{$\pm$} \scriptsize{1.90} & 54.53 \scriptsize{$\pm$} \scriptsize{2.66} & 41.72 \scriptsize{$\pm$} \scriptsize{2.88} & \textbf{61.82} \scriptsize{$\pm$} \scriptsize{4.34} \\

$0.05$ & 58.74 \scriptsize{$\pm$} \scriptsize{1.28} & 68.83 \scriptsize{$\pm$} \scriptsize{2.70} & 67.42 \scriptsize{$\pm$} \scriptsize{2.69} & 58.17 \scriptsize{$\pm$} \scriptsize{2.07} & \textbf{74.61} \scriptsize{$\pm$} \scriptsize{2.72} \\
\bottomrule
\end{tabular}

\caption{Experimental Results on the reduced Colored MNIST dataset}
\label{table:cmnist}
\end{table*}

\begin{figure*}
\begin{center}
\includegraphics[width=0.9\textwidth]{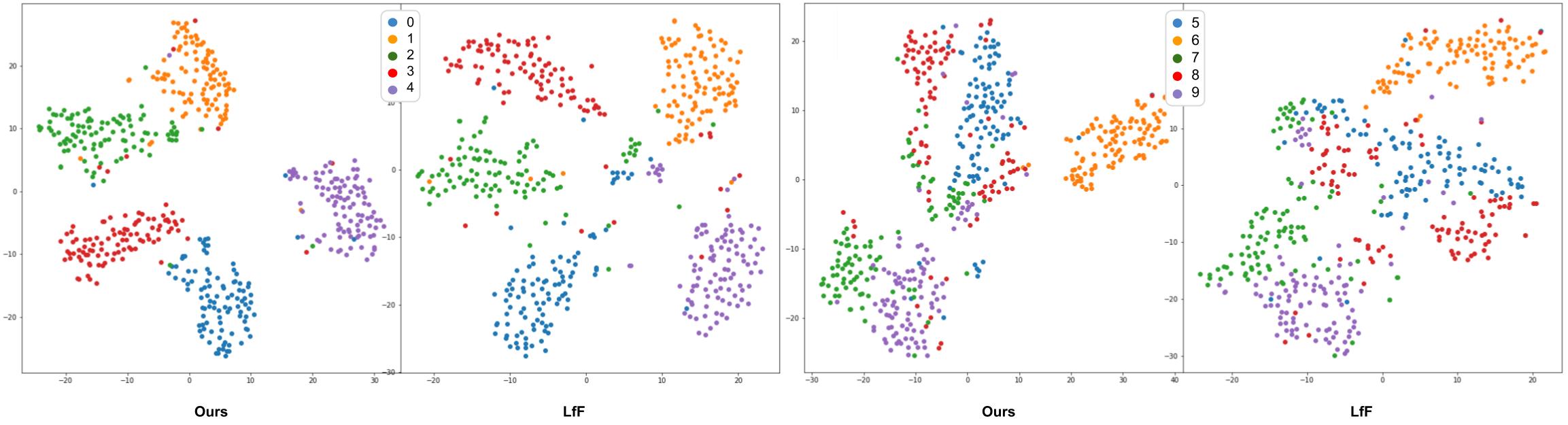}
\end{center}
\vspace{-2pt}
   \caption{t-SNE Plot for the features extracted by the model trained using LfF and our approach on the reduced Colored MNIST Dataset.}
   \vspace{-5pt}
\label{fig:tsne}
\end{figure*}

\subsection{Results on Colored MNIST}

The results of the experiments conducted on the reduced Colored MNIST dataset with bias-conflicting samples ratio of 0.02 and 0.05 are presented in Table \ref{table:cmnist}. The Vanilla approach, which represents a simple MLP model with no debiasing approach applied, achieves an accuracy of 39.47\%, 43.66\% and 58.74\% for $\sigma$ equal to 0.01, 0.02 and 0.05, respectively. Applying the DebiAN~\cite{debian} method significantly decreases the model performance even though it is a recent state-of-the-art method when trained on the full dataset. Specifically, the performance of DebiAN is lower than the vanilla model by absolute margins of 3.61\% and 1.94\% for $\sigma=0.01$ and 0.02, respectively, and almost the same as the vanilla model for $\sigma=0.05$. Since DebiAN identifies bias and removes it alternatingly, it most likely is unable to effectively identify the bias with the extremely limited number of bias-conflicting samples in this setting. 

The results indicate that the proposed approach outperforms the vanilla method by absolute margins of 9.45\%, 18.16\%, and 15.87\% for the $\sigma = 0.01, 0.02$, and 0.05, respectively. Further, the proposed approach outperforms the closest method LfF by absolute margins of 5.26\%, 5.24\%, and 5.78\%, for the $\sigma = 0.01, 0.02$ and 0.05, respectively. The proposed approach also outperforms LDD~\cite{ldd} by absolute margins of 6.58\%, 7.29\%, and 7.19\%, for the $\sigma = 0.01, 0.02$ and 0.05, respectively. The LfF method relies heavily on reweighting the available bias-conflicting samples, which leads to overfitting since the bias-conflicting samples are very limited in this setting. LDD also suffers from overfitting for the same reason and, therefore, the feature level augmentations used by LDD are not as effective in the limited data setting as compared to the full data setting. In contrast, our proposed approach identifies likely bias-conflicting samples and synthesizes hybrid samples while maintaining diversity in such samples. As a result, the population of bias-conflicting samples increases and is also diverse, which reduces overfitting and improves debiasing. We also compare the features extracted by the debiased model trained using LfF and our approach on the reduced Colored MNIST dataset ($\sigma=0.05$) using a t-SNE plot (see Fig.~\ref{fig:tsne}). The plot clearly indicates that the features extracted by the debiased model trained using our approach are better clustered according to their respective classes as compared to LfF. Therefore, our approach leads to a more discriminative feature space, which improves the model performance.

\subsection{Results on Corrupted CIFAR-10 Type 0 }

Table \ref{table:cifar0} shows the results of our experiments on the Corrupted CIFAR-10 Type 0 dataset. The vanilla model without any debiasing approach achieves a very low accuracy of 25\%, 25.68\%, and 29.40\% for $\sigma=0.01, 0.02$, and 0.05, respectively. The results indicate that LfF \cite{lff} fails to significantly improve the model performance as compared to the vanilla model. Specifically, LfF is able to improve upon the vanilla model by a very small absolute margin of 0.44\% for $\sigma=0.01$ and performs very similar to the vanilla model for $\sigma=0.02$ and 0.05. The LDD \cite{ldd} approach also fails to outperform the vanilla model significantly. The state-of-the-art DebiAN~\cite{debian} suffers significantly in this setting and even negatively affects the model performance in some cases. In contrast, the proposed approach outperforms LfF by absolute margins of 5.22\%, 4.20\%, and 5.59\%, and outperforms LDD by absolute margins of 3.24\%, 4.39\% and 4.07\%, for $\sigma=0.01, 0.02$, and 0.05, respectively. 
\begin{table*}[t]
\centering
\begin{tabular}{cccccc}
\toprule
$\sigma$ & Vanilla & LfF & LDD & DebiAN & Ours \\
\midrule
$0.01$ &  25.00 \scriptsize{$\pm$} \scriptsize{0.35} & 25.44 \scriptsize{$\pm$} \scriptsize{0.41} & 27.42 \scriptsize{$\pm$} \scriptsize{0.96} & 24.66 \scriptsize{$\pm$} \scriptsize{0.49} & \textbf{30.66} \scriptsize{$\pm$} \scriptsize{1.28} \\
$0.02$ & 25.68 \scriptsize{$\pm$} \scriptsize{0.82} & 27.53 \scriptsize{$\pm$} \scriptsize{0.65} & 27.34 \scriptsize{$\pm$} \scriptsize{0.65} & 26.00 \scriptsize{$\pm$} \scriptsize{0.15} & \textbf{31.73} \scriptsize{$\pm$} \scriptsize{0.32} \\
$0.05$ & 29.40 \scriptsize{$\pm$} \scriptsize{0.65} & 30.71 \scriptsize{$\pm$} \scriptsize{0.98} & 32.23 \scriptsize{$\pm$} \scriptsize{0.47} & 29.01 \scriptsize{$\pm$} \scriptsize{0.23} & \textbf{36.30} \scriptsize{$\pm$} \scriptsize{1.27} \\
\bottomrule 
\end{tabular}
\caption{Experimental Results on the reduced Corrupted CIFAR-10 Type 0 dataset}
\vspace{-5pt}
\label{table:cifar0}
\end{table*}
\begin{table*}[t]
\centering
\begin{tabular}{cccccc}
\toprule
$\sigma$ & Vanilla & LfF & LDD & DebiAN & Ours \\
\midrule
$0.01$ &  25.03 \scriptsize{$\pm$} \scriptsize{0.69} & 25.33 \scriptsize{$\pm$} \scriptsize{1.12} & 24.68 \scriptsize{$\pm$} \scriptsize{1.52} & 25.48 \scriptsize{$\pm$} \scriptsize{0.12} & \textbf{31.69} \scriptsize{$\pm$} \scriptsize{0.79} \\
$0.02$ & 26.73 \scriptsize{$\pm$} \scriptsize{0.47} & 27.90 \scriptsize{$\pm$} \scriptsize{0.87} & 27.93 \scriptsize{$\pm$} \scriptsize{2.24} & 27.04 \scriptsize{$\pm$} \scriptsize{0.32} & \textbf{34.42} \scriptsize{$\pm$} \scriptsize{0.70} \\

$0.05$ & 30.84 \scriptsize{$\pm$} \scriptsize{0.69} & 32.15 \scriptsize{$\pm$} \scriptsize{1.60} & 33.16 \scriptsize{$\pm$} \scriptsize{3.45} & 31.56 \scriptsize{$\pm$} \scriptsize{0.81} & \textbf{41.23} \scriptsize{$\pm$} \scriptsize{0.60} \\
\bottomrule
\end{tabular}
\caption{Experimental Results on the reduced Corrupted CIFAR-10 Type 1 dataset}
\vspace{-5pt}
\label{table:cifar1}
\end{table*}

\begin{table}[t]
\centering
 \scalebox{0.8}{
 \addtolength{\tabcolsep}{-3pt}
\begin{tabular}{cccccc}
\toprule
$\sigma$ & Vanilla & LfF & LDD & DebiAN & Ours \\
\midrule
$0.05$ & 59.86 \scriptsize{$\pm$} \scriptsize{0.57} & 59.20 \scriptsize{$\pm$} \scriptsize{3.02} & 59.13 \scriptsize{$\pm$} \scriptsize{1.33} & 58.40 \scriptsize{$\pm$} \scriptsize{0.99} & \textbf{63.86} \scriptsize{$\pm$} \scriptsize{1.03} \\
\bottomrule
\end{tabular}
}
\caption{Experimental Results on the reduced BFFHQ dataset}
\label{table:bffhq}
\vspace{5pt}
\end{table}

\subsection{Results on Corrupted CIFAR-10 Type 1}
The results of our experiments on Corrupted CIFAR-10 Type 1 are reported in Table~\ref{table:cifar1}. The vanilla model achieves a very poor performance of 25.03\%, 26.73\%, and 30.84\% accuracy for $\sigma=0.01,0.02$ and $0.05$, respectively. LfF is not able to significantly outperform the vanilla model, achieving a performance of 25.33\%, 27.90\% and 32.15\% accuracy for $\sigma=0.01, 0.02$, and 0.05, respectively. LDD achieves slightly better performance than LfF in some cases. The DebiAN method performs very close to the vanilla model and is not very effective. The proposed approach significantly outperforms the vanilla, LfF, LDD, and DebiAN methods by absolute margins of 10.39\%, 9.08\%, 8.07\%, and 9.67\% for $\sigma$ = 0.05.

\begin{table*}[t]
    \centering
\begin{minipage}{0.63\textwidth}
     
                \centering

\includegraphics[width=0.9\textwidth]{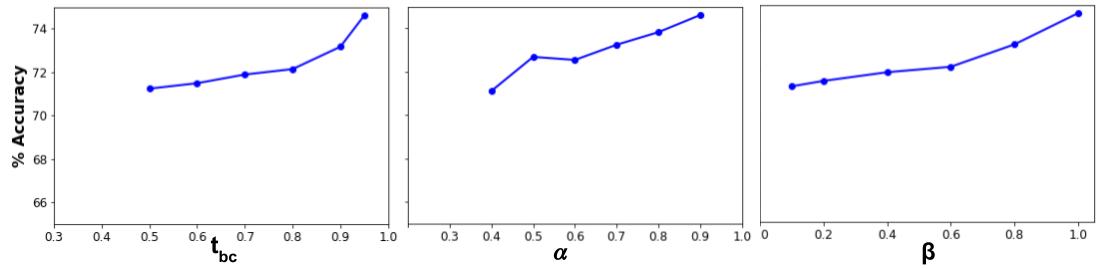}

   \captionof{figure}{\small{Choice of hyperparameters for the reduced Colored MNIST dataset}}
\label{fig:hypermnist}
                \includegraphics[width=0.9\textwidth]{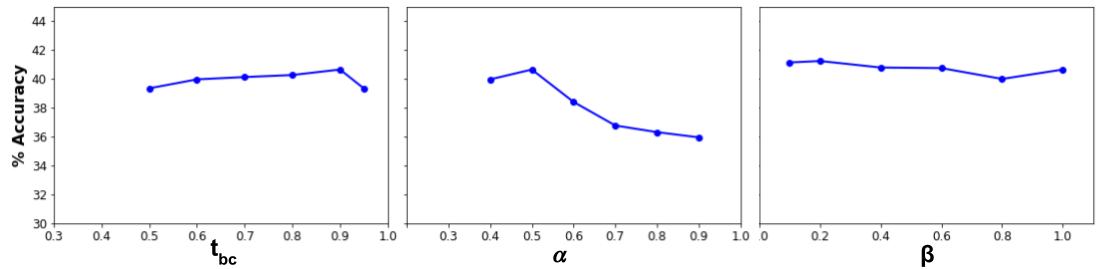}
        \captionof{figure}{Hyperparameters selection for the reduced Corrupted CIFAR-10 dataset}\label{fig:hyper}

\includegraphics[width=0.9\textwidth]{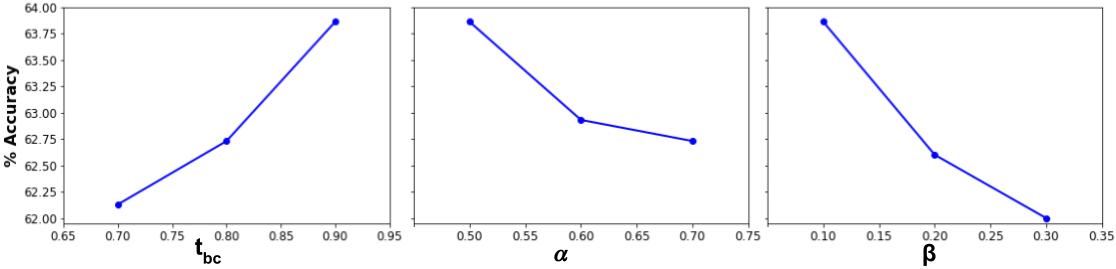}

     \captionof{figure}{\small{Choice of hyperparameters for the reduced BFFHQ dataset}}
\label{fig:hyper_bffhq}
\vspace{-5pt}
                
\end{minipage}  
    \hfill
\begin{minipage}{0.35\textwidth}
\includegraphics[width=\textwidth]{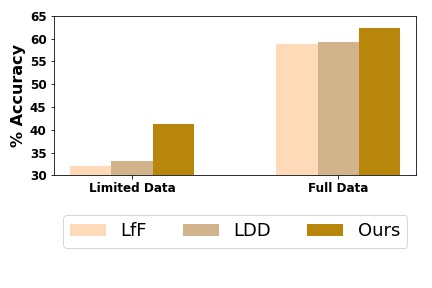}
                 \captionof{figure}{Experimental results on Corrupted CIFAR-10 Type 1 with limited and full data.}
                   \label{fig:limitedfullbar}

\centering
\vspace{10pt}
\begin{tabular}{ccc}
\hline
Mixup & CutMix & Ours \\
\hline
53.38 & 58.88 & 74.61\\
\hline
\end{tabular}

\captionof{table}{Results on reduced Colored MNIST with $\sigma$=0.05 for different strategies of synthesizing hybrid samples}
\label{table:mixup}

\end{minipage} 
\end{table*}

\subsection{Results on BFFHQ}

The experimental results on the BFFHQ dataset are reported in Table~\ref{table:bffhq}. Similar to the other datasets, the results indicate that the performance of existing debiasing approaches is very poor when the training data is limited. In fact, on BFFHQ, the compared debiasing approaches perform even worse than the vanilla baseline method. In contrast, the proposed approach outperforms the vanilla, LfF, LDD, and DebiAN approaches by absolute margins of 4.00\%, 4.66\%, 4.73\%, and 5.46\%, respectively.

\subsection{Ablation Experiments}

\subsubsection{Hyperparameter Selection}
We report the results of the experiments to identify the suitable values for the hyperparameters $t_{bc}$, $\alpha$, $\beta$, for the reduced Colored MNIST, the reduced Corrupted CIFAR-10 and the reduced BFFHQ datasets in Figs.~\ref{fig:hypermnist},~\ref{fig:hyper},~\ref{fig:hyper_bffhq}, respectively. We choose the best values for each hyperparameter.

\subsubsection{Analysis of other Performance Aspects as Training Progresses} 

\begin{figure}
\begin{center}
\begin{subfigure}[b]{.42\textwidth}
  \centering
  \includegraphics[width=\linewidth]{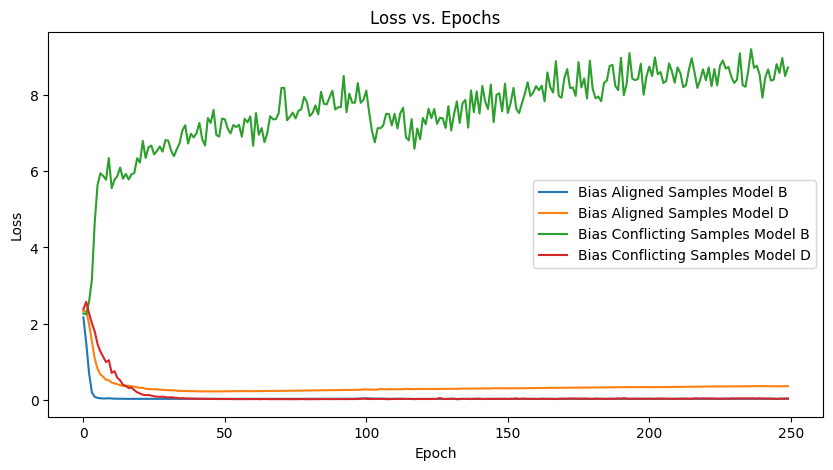}
  \caption{}
  
\end{subfigure}
\hfill
\begin{subfigure}[b]{.42\textwidth}
  \centering
  \includegraphics[width=\linewidth]{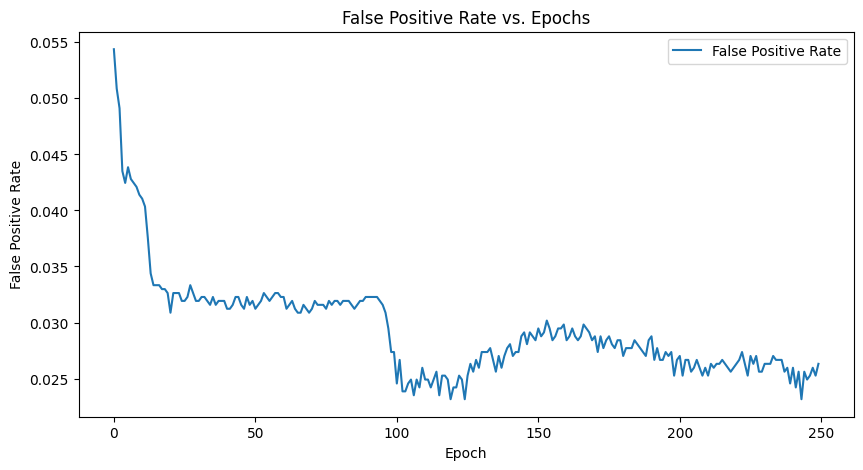}
  \caption{}
  
\end{subfigure}
\end{center}
\vspace{-20pt}
   \caption{Analysis of other performance aspects. a) Change in training loss for unbiased (bias-conflicting) and biased (bias-aligned) samples b) Change in the false positive rate of the likely bias-conflicting samples identified using our approach}
   \vspace{-5pt}
\label{fig:plot}
\end{figure}

Fig.~\ref{fig:plot} (a), shows the training loss incurred by $M_D$ and $M_B$ models for the bias-aligned and bias-conflicting samples on the reduced colored MNIST dataset with $\sigma=0.05$. We observe that as the training progresses, the training loss of $M_B$ for the bias-conflicting samples increases and remains extremely high as compared to the bias-aligned samples. This is expected since we are training $M_B$ to be biased. In contrast, the training loss of $M_D$ is comparatively low for both bias-aligned and bias-conflicting samples. Fig.~\ref{fig:plot} (b), shows the false positive rate of likely bias-conflicting samples identified using our method. We observe that the false positive rate keeps dropping as the training progresses and remains extremely low. This demonstrates the effectiveness of our approach.

\subsubsection{Comparison of Strategies for Synthesizing Hybrid Samples in our Approach}

In the proposed approach, the likely bias-conflicting samples have to be combined with likely bias-aligned samples to synthesize hybrid samples. We perform experiments to compare different strategies for doing so, including our proposed strategy. The results in Table~\ref{table:mixup} indicate that the samples produced through our approach are significantly better at improving the model performance. Also note that for the compared mixing methods, we follow the methods given in the original papers, which includes label mixing, which is not needed in our approach.

\subsubsection{Performance degradation of Debiasing Methods in Limited Data Regime}\label{sec:abldata}
LfF~\cite{lff} and LDD~\cite{ldd} are popular debiasing approaches that are able to achieve only 32.15\% and 33.16\% on the Corrupted CIFAR-10 Type 1 dataset when only 10\% of the training data is present with a bias-conflicting ratio of 0.05. This is very close to the simple baseline vanilla model, which achieves 30.84\%. This is in stark contrast to the experiments where the full Corrupted CIFAR-10 Type 1 dataset is used, and the baseline, LfF, and LDD achieve 41.27\%, 58.77\%, and 59.3\% accuracy, respectively, in that setting(see Fig.~\ref{fig:limitedfullbar}). This indicates that in a limited data regime, the effect of biases on the model predictions is extremely high, and the effectiveness of debiasing approaches is significantly degraded. The results also show that our approach is significantly more effective at debiasing deep learning models in the limited data settings as compared to other debiasing approaches.

\subsubsection{Performance on Sufficient Training Data}\label{sec:fulldata}
We experimentally verify the performance of the proposed method when the entire Corrupted CIFAR-10 Type 1 data is present with a bias-conflicting ratio of 0.05. We observe that our approach achieves 62.3\% accuracy in the presence of sufficient training data and outperforms LfF and LDD by absolute margins of 3.53\% and 3\%, respectively (see Fig.~\ref{fig:limitedfullbar}). The results also indicate that the drop in performance of our approach due to limited data is significantly less as compared to the other debiasing approaches.

\subsubsection{Performance on Sufficient Training Data and Very Low Bias} \label{sec:fulldatalowbias}
We performed experiments to verify whether our debiasing approach has any negative effect when the data is sufficient and has very low bis. Specifically, we experiment on the entire Corrupted CIFAR-10 Type 1 dataset, with p=1 and high $\sigma$, i.e., 0.4, 0.5, such that the data is sufficient and highly balanced, therefore, suffering from very low bias. The results in Table~\ref{table:balanced} indicate that when the bias-conflicting and aligned samples are in balance, the vanilla model performs much better, indicating a low effect of bias, but the LfF debiasing method significantly hurts the performance in the process of debiasing the network. In contrast, the proposed approach does not have any negative effects and, in fact, improves model performance even in this setting.

\begin{table}[t]
\centering
 \scalebox{1}{
\begin{tabular}{cccccc}
\hline
$\sigma$ & Vanilla & LfF & LDD & DebiAN & Ours \\
\hline
$0.4$ & 72.99  & 71.26 & 73.72 & 72.57 & 73.93\\
$0.5$ & 74.64  & 71.79 & 75.05 & 75.01 & 75.12\\
\hline
\end{tabular}
}
\caption{Results on Corrupted CIFAR-10 Type 1 for full data ($p=1$) with a good balance between bias-aligned and bias-conflicting samples, i.e., very low bias}
\vspace{-10pt}
\label{table:balanced}
\end{table}

\subsubsection{Alternative to Filtering Ratio}\label{sec:medianmean}
In the proposed approach, we select the top $N*(1-t_{bc})$ samples in a batch of size $N$ with the highest reweighting factor as the likely bias-conflicting samples. Another alternative is to identify a threshold for the reweighting factor and consider all samples with a reweighting factor above a threshold as the likely bias-conflicting samples. We experimented with using the mean reweighting factor in a batch as the threshold and also using 0.1, 0.3, 0.5, 0.7, and, 0.9 as the threshold. However, we experimentally observe that using the filtering ratio instead of threshold outperforms the above options by absolute margins of 3.17\%, 2.37\%, 1.21\%, 0.92\%, 3.39\%, 9.04\%, respectively, on the Corrupted CIFAR-10 Type 1 dataset ($\sigma=0.05$).

\section{Conclusion}
In this paper, we have proposed a novel approach to address the problem of biased predictions in the limited data setting. Existing debiasing approaches suffer from a massive performance drop in the limited data setting due to the increased influence of the more populous bias-aligned samples in this setting and their additional focus on the bias-conflicting samples, which are extremely limited in this setting, causing overfitting. The proposed approach identifies likely bias-conflicting samples and synthesizes diverse hybrid samples from them to better handle the problems in this setting. We experimentally demonstrated that our approach performs significantly better than existing debiasing techniques in the limited data setting.

\section{Acknowledgements}
\noindent
Assistance from DST/INSPIRE/04/2021/003225 and IIT Jodhpur is duly acknowledged.
{\small
\bibliographystyle{ieee_fullname}
\bibliography{egbib}
}

\section{Appendix}

\subsection{Datasets and Additional Implementation Details}
We utilized the following datasets in our experiments: Colored MNIST, Corrupted CIFAR-10, and BFFHQ.  

Colored MNIST \cite{lff} is a variant of the original MNIST dataset \cite{mnist} with color bias. A specific color is injected into the images of the MNIST dataset with some perturbation. Each digit in Colored MNIST is associated with a specific foreground color. Consequently, the naive baseline models learn to classify some digits based on only the color of the digit, which is an unwanted correlation/bias that such models learn.
    
Corrupted CIFAR-10, on the other hand, consists of ten different types of texture biases applied to the CIFAR-10 dataset \cite{cifar}. This dataset was constructed following the design protocol of Hendrycks and Dietterich \cite{hendrycks2019benchmarking}, and each class is highly correlated with a particular texture. Specifically, Corrupted CIFAR-10 Type 0 contains texture biases such as snow, frost, fog, brightness, contrast, spatter, elastic, JPEG, pixelate, and saturate, while Corrupted CIFAR-10 Type 1 contains texture biases such as Gaussian noise, shot noise, impulse noise, speckle noise, Gaussian blur, defocus blur, glass blur, motion blur, zoom blur, and original.

The Biased FFHQ (BFFHQ) dataset was derived from the FFHQ dataset \cite{ffhq} by the authors of \cite{ldd}, which consists of human face images labeled with various facial attributes. The BFFHQ comprises age and gender as the intrinsic and biased attributes, respectively, and compiles a collection of images that exhibit a strong correlation between these two attributes. In order to achieve this bias, a majority of the images of males have males with ages between 40 and 59, and a majority of the images of females have females with ages between 10 and 29. Therefore, a majority of the females in the dataset are young, while a majority of the males in the dataset are old. The bias-aligned samples of the dataset are the samples that follow this bias.

In this work, we use the PyTorch framework \cite{paszke2017automatic} and Python 3.0 for all the experiments. We use the NVIDIA RTX A5000 graphics processing unit for our experiments. We run all the experiments 3 times with different seeds and report the average accuracy and the 95\% confidence score.  

\subsection{Additional Experiments with p = 1\% and 2\%}

\begin{table}[h]
\centering
 \scalebox{1}{

\begin{tabular}{cccccc}

\hline
$p$ & Vanilla & LfF & LDD & DebiAN & Ours \\

\hline
$2\%$ & 57.04  & 65.58 & 63.18 & 56.54 & 69.08\\

$1\%$ & 47.51  & 52.33 & 51.02 & 47.43 & 55.72\\
\hline

\end{tabular}
}

\caption{Additional Experimental Results on reduced CMNIST for $\sigma=0.05$}

\label{table:pvalue}
\end{table}

We perform additional experiments with a significantly limited amount of training data, i.e., p=1\% and 2\% of the training data. The results for these experiments on the reduced Colored MNIST dataset with $\sigma=0.05$ are reported in Table~\ref{table:pvalue}. The results indicate that the performance of the vanilla model falls significantly as compared to the results in the main paper as the amount of training data decreases further. The results indicate that the effectiveness of the bias mitigation approaches decreases even further. The results also indicate that the proposed approach significantly outperforms the vanilla, LfF, LDD, and DebiAN approaches by absolute margins of 12.04\%, 3.5\%, 5.9\%, 12.54\%, respectively, for p = 2\% and by absolute margins of 8.21\%, 3.39\%, 4.7\%, 8.29\%, respectively, for p=1\%. Therefore, the proposed approach is more effective at debiasing models as compared to the other approaches in this setting.

\end{document}